\documentclass[oneside,tablecaption=bottom,wcp]{jmlr} 
 
\usepackage{layouts}
\usepackage{arydshln}

\usepackage{microtype}
\usepackage{caption}
\usepackage{booktabs} 

\usepackage{enumitem}

\setenumerate[1]{align=left,label=\arabic*}

\usepackage{cancel}

\SetKwInput{KwData}{Inputs}
\SetKwInput{KwResult}{Output}

\usepackage[utf8]{inputenc} 
\usepackage[T1]{fontenc}    
\usepackage{amsfonts}       
\usepackage{nicefrac}       
\usepackage{floatrow}

\usepackage{mathtools}
\usepackage{bm, bbm}

\usepackage{wrapfig}

\usepackage{nth}
\usepackage{nicefrac}

\usepackage{cleveref} 
\crefformat{equation}{(#2#1#3)}
\crefformat{figure}{Figure #2#1#3}
\crefformat{table}{Table #2#1#3}
\crefformat{appendix}{Appendix #2#1#3}
\crefformat{section}{Section #2#1#3}

\usepackage{multirow}
\usepackage{caption}
\usepackage{enumitem}

\usepackage{float}

\usepackage{pgf}
\usepackage{pgfplots}
\usepackage{blindtext}
\newfloatcommand{capbtabbox}{table}[][\FBwidth]

\renewcommand{\det}[0]{\text{det}}

\newcommand{\RN}[1]{%
  \textup{\uppercase\expandafter{\romannumeral#1}}%
}

\usepackage{amsmath,amsfonts,bm}

\def\eqref#1{equation~\ref{#1}}

\def\1{\bm{1}}

\DeclareMathAlphabet{\mathsfit}{\encodingdefault}{\sfdefault}{m}{sl}
\SetMathAlphabet{\mathsfit}{bold}{\encodingdefault}{\sfdefault}{bx}{n}

\def\gH{{\mathcal{H}}}

\def\gL{{\mathcal{L}}}
\def\gM{{\mathcal{M}}}
\def\gN{{\mathcal{N}}}

\newcommand{\E}{\mathbb{E}}

\newcommand{\R}{\mathbb{R}}

\DeclareMathOperator*{\argmax}{arg\,max}
\DeclareMathOperator*{\argmin}{arg\,min}

\jmlrproceedings{AABI 2023}{5th Symposium on Advances in Approximate Bayesian Inference, 2023}

\title[Online Laplace Model Selection Revisited]{Online Laplace Model Selection Revisited}

\author{
\Name{Jihao Andreas Lin}{\normalfont\textsuperscript{1,2}}
\Email{jal232@cam.ac.uk} \\
\Name{Javier Antorán}{\normalfont\textsuperscript{1}}
\Email{ja666@cam.ac.uk} \\
\Name{José Miguel Hernández-Lobato}{\normalfont\textsuperscript{1}}
\Email{jmh233@cam.ac.uk} \\
\addr \textsuperscript{1}University of Cambridge \quad \textsuperscript{2}Max Planck Institute for Intelligent Systems}

\begin{document}

\maketitle

\begin{abstract}
The Laplace approximation provides a closed form model selection objective for neural networks (NN).
\emph{Online} variants, which optimise NN parameters jointly with hyperparameters, like weight decay strength, have seen renewed interest in the Bayesian deep learning community.
However, these methods violate Laplace's method's critical assumption that the approximation is performed around a mode of the loss, calling into question their soundness. 
This work re-derives online Laplace methods, showing them to target a variational bound on a variant of the Laplace evidence which does not make stationarity assumptions, i.e, a mode-corrected method. 
Online Laplace and its mode-corrected counterpart share stationary points where 1. the NN parameters are a \emph{maximum a posteriori}, satisfying Laplace's method's assumption, and 2. the hyperparameters maximise the Laplace evidence, motivating online methods. 
We demonstrate that these optima are roughly attained in practise by online algorithms using full-batch gradient descent on UCI regression datasets. Here, online model selection prevents overfitting and outperforms validation-based early~stopping.
\end{abstract}

\section{Introduction}

Online model selection holds the promise of automatically tuning large numbers of neural network (NN) hyperparameters during a single training run, obsoleting cross validation \citep{clarke2022scalable}.
Online Laplace methods (OL) interleave regular NN optimisation steps with steps of hyperparameter optimisation with respect to the Laplace-approximated model evidence \citep{Foresee97BayesNewton,Friston07variational,Immer2021Marginal}. The method has been used to learn layer-wise weight decay strengths, data noise scales, and even data augmentation hyperparameters \citep{immer2022invariance}.

Laplace's method constructs a second-order Taylor expansion around a mode of the posterior. Here, the first-order term vanishes, leaving a pure quadratic approximation to the log-density, that is, a Gaussian. The Laplace evidence is this distribution's normalising constant. 
The online setting implies non-convergence of the NN parameters, violating the stationarity assumption. \cite{antoran2022Adapting} show that applying Laplace's method to pre-trained NNs which do not attain zero loss gradient leads to severe deterioration in model selection performance.
\cite{Immer2021Marginal} experiment with inclusion of the first-order Taylor term in the OL procedure but find that it leads to training instability.

This work revisits online Laplace methods, taking steps towards reconciling their seeming unsoundness with their satisfactory empirical performance.

\begin{itemize}
    \item We show that a variant of OL that includes the first-order Taylor term (and thus does not assume stationarity) corresponds to NN hyperparameter optimisation with the evidence of a tangent linear model. We re-derive the standard OL objective as a variational lower bound on the evidence of this tangent linear model, motivating~its~use.
    
    \item  We show OL shares fixed points, where the ELBO is tight, with the above first-order-corrected procedure. Here, the linear model's \emph{maximum a posteriori} (MAP) parameters match the NN parameters, for which they are also a MAP, and the linear model's evidence matches the NN's Laplace evidence, further motivating its use.
    
    \item  We show that these fixed points are roughly attained in practise by small NNs trained via full-batch gradient descent with OL hyperparameter tuning on UCI regression datasets, but they are not attained without online tuning.
    The hyperparameters found through online Laplace lead to NNs that outperform their non-online counterparts.
\end{itemize}

\section{Preliminaries: Laplace approximation and online variants}

We consider a supervised problem where $n$ inputs, stacked as $X \in \R^{n \times d_x}$, are paired with scalar targets stacked as $y \in \R^{n}$. We introduce a NN regressor $f: \R^{d_w} \times \R^{d_x} \to \R$ with $w \in \R^{d_w}$ as parameters. Omitting dependence on the inputs from our notation, as these are fixed, and working with arrays of stacked predictions $f(w) \in \R^{n}$, we model the targets as
\begin{gather*} \label{eq:NN_problem_setup}
 y = f(w) + \epsilon \quad \textrm{with}  \quad w \sim \gN(0, \alpha^{-1} I_{d_w}) \quad \textrm{and} \quad \epsilon \sim \gN(0, \beta^{-1} I_n),
\end{gather*}
where $\alpha$ is the scalar precision of our prior over NN parameters and $\beta$ is the scalar homoscedastic noise precision. The set of both hyperparameters is $\gM = \{\alpha, \beta\}$. Our results readily generalise to vector-valued hyperparameters and outputs \citep{antoran2023samplingbased}.

We train the NN to minimise the loss $\ell_f(w; \gM) = \frac{\beta}{2} \|y - f(w)\|^2 + \frac{\alpha}{2} \|w\|^2$.
This objective matches the unnormalised, negative log-joint density of our parameters and targets  
\begin{gather*}
    e^{-\ell_f(w; \gM)} \left(\frac{\beta}{2 \pi} \right)^{\frac{n}{2}} \left(\frac{\alpha}{2 \pi}\right)^{\frac{d_w}{2}}
    = \gN(y;\, f(w), \beta^{-1} I_n)  \gN(w;\, 0, \alpha^{-1} I_{d_w})
    \coloneqq p_f(y, w; \gM).
\end{gather*} 
Laplace's method \citep{Mackay1992Thesis, bishop2006pattern} takes a second-order Taylor expansion of $\log p_f(y, w; \gM)$ around $w_0$, assumed to be a mode $(w_0 \in \argmin_w \ell_f(w; \gM))$, and approximates the NN model evidence $p_f(y; \gM)$ by integrating the expanded expression as
\begin{align} \label{eq:standard_laplace_evidence}
& \int p_f(y, w; \gM) \; \mathrm{d} w  \approx \int \exp \left(
    \log  p_f(y, w_0; \gM)
    -\frac{1}{2}(w - w_0)^\top
    H
    (w - w_0)
    \right) \; \mathrm{d} w \notag  \\
    &=  \exp \left( \frac{d_w}{2} \log \alpha
    + \frac{n}{2} \log \beta
    - \frac{\beta}{2} \lVert y {-} f(w_0) \rVert_2^2
    {-} \frac{\alpha}{2} \lVert w_0 \rVert_2^2
    - \frac{1}{2}\log \det H
    - \frac{n}{2} \log 2\pi \right) \\
    & \coloneqq \exp \gL_f(\gM; w_0) \notag, 
\end{align}
where the first-order term vanishes since we assume $\partial_{w}\ell_f(w_0; \gM) = 0$ and $H$ is $\partial^2_w \ell_f(w_0; \gM)$. 
In practice, the loss Hessian is replaced by the Generalised Gauss Newton matrix (GGN), that is, $H = \beta J^T J + \alpha I_{d_w} \in \R^{d_w \times d_w}$ for $J =\partial_w f(w_0) \in \R^{n \times d_w}$ the NN's Jacobian evaluated at the training inputs.
This choice provides ease of computation and numerical stability \citep{Martens2020insights,Immer2021Improving}.
We hereon assume $H$ to be the GGN and suppress its dependence on the hyperparameters and the expansion point from our notation.

\subsection{Online hyperparameter tuning}\label{subsec:preliminaries_online}

We may tune the hyperparameters $\gM$ for a NN by choosing them to maximise $\gL_f(\gM; w^\star)$, where $w^\star \in \argmin_w \ell_f(w; \gM)$ is a true optima of the loss. This will not change our NN's predictions however, as its parameters are held fixed at $w_\star$. \cite{Mackay1992Thesis} proposes to re-train the NN from scratch using the updated hyperparameters. This procedure terminates when a joint stationary point of the parameters and hyperparameters is found:
$w_\star \in \argmin_w \ell_f(w; \gM^\star)$ and $\gM^\star \in \argmin_\gM \gL_f(\gM; w^\star)$.

\begin{wrapfigure}{r}{.4\textwidth}
\vspace{-15pt}
\centering
\includegraphics[width=.8\textwidth]{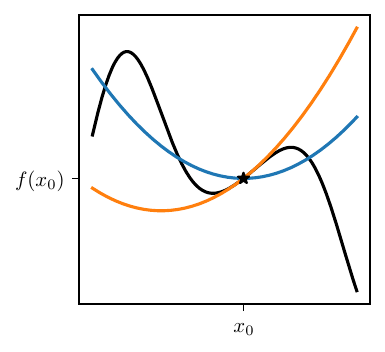}
\vspace{15pt}
\caption{Second-order Taylor approximations of a function $f$ (black) around $x_0$ (star), with (orange) and without (blue) the first-order term.}
\label{fig:illustration}
\vspace{-30pt}
\end{wrapfigure}
The size of NN parameter spaces and datasets has grown dramatically since 1992 and re-training the NN is now impractical. 
This motivates online Laplace (OL) approaches which, at timestep $t$ with parameters $w_t$, perform a step of NN parameter optimisation to minimise $\ell_f(w_t; \gM)$, followed by a hyperparameter update to maximise $\gL_f(\gM; w_t)$ \citep{ Foresee97BayesNewton,Friston07variational,Immer2021Marginal}\footnote{\cite{Friston07variational} refers to the described online Laplace procedure as \emph{Variational Laplace}.}.
Importantly, the Taylor expansion point is chosen to match the current NN parameter setting $w_t$.
Since optimisation has not converged, $w_t \notin \argmin_w \ell_f(w; \gM)$, then $\gL_f(\gM; w_t)$, which discards the first-order expansion term, \emph{does not provide a local approximation to the evidence.} \Cref{fig:illustration} illustrates that discarding this first-order term can result in a vastly different~approximation.




\section{Understanding online Laplace through the tangent model}\label{sec:contributions}

This section re-derives online Laplace methods for NNs as empirical Bayesian inference in a tangent linear model. This analysis provides a non-heuristic motivation for these methods. 

\paragraph{Online hyperparameter optimisation including the first-order term}
To start, we integrate the Taylor expansion of $\log p_f(y, w; \gM)$ at $w_t$ without discarding the first-order~term
\begin{align} \label{eq:corrected_laplace_evidence}
&\int {\exp} \left(
    \log  p_f(y, w_t; \gM) - \partial_w \ell_f(w_t; \gM) (w - w_t)
    -\frac{1}{2}(w - w_t)^\top H
    (w - w_t)
    \right) \mathrm{d} w \\
    &= {\exp}\left( \frac{d_w}{2} \log \alpha
    {\,+\,} \frac{n}{2} \log \beta
    {\,-\,} \frac{\beta}{2} \lVert y {-} (f(w_t) {+} J(v^\star {-} w_t)) \rVert_2^2
    {\,-\,} \frac{\alpha}{2} \lVert v^\star \rVert_2^2
    {\,-\,} \frac{1}{2}\log \det H
    {-} \frac{n}{2} \log 2\pi \right) \notag \\
    &\coloneqq \exp \gL_h(\gM; w_t), \notag
\end{align}
where the optimum of the quadratic expansion $v^\star$ is obtained by performing a Gauss-Newton update on the NN loss at $w_t$, that is, $v^\star = w_t - \beta H^{-1}\left(J^T(f(w_t) - y) + \alpha w_t\right)$. Gauss-Newton optimisation is common in the NN second order optimisation-literature; it makes more progress per step than gradient descent \citep{botev2017Optimisation,Martens2020insights,barbano2023fast}. Since $\gL_h(\gM; w_t)$ represents a local evidence approximation, mode-corrected by a Gauss-Newton step, it may more closely match the Laplace evidence computed at an optimum $\gL_f(\gM; w^\star)$ than the Laplace evidence computed at the online expansion point~$\gL_f(\gM; w_t)$.
We may use $\gL_h(\gM; w_t)$ to construct an algorithm where each step of NN parameter optimisation minimising $\ell_f(w_t; \gM)$ is interleaved with a hyperparameter update maximising $\gL_h( \gM; w_t)$.


\begin{figure}[t]
\vspace{-6pt}
\centering
\includegraphics[width=0.95\textwidth]{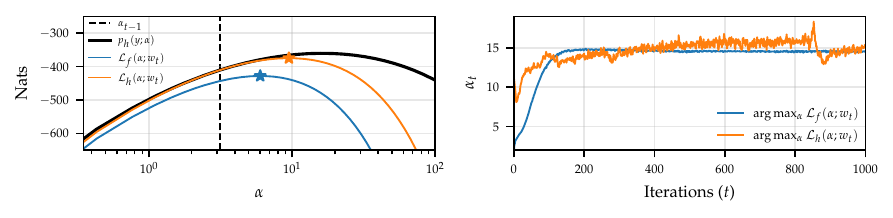}
\vspace{-6pt}
\caption{Illustration of exact linear model evidence, and LM and OL hyperparameter objectives at a single train step $t$ (left). The latter two represent lower bounds as per \cref{eq:amazing_bound}. $\mathcal{L}_h$ is tighter which leads to larger updates, making hyperparameter trajectories unstable (right).}
\label{fig:ELBO_alpha_trace}
\vspace{-7pt}
\end{figure}

\paragraph{Online hyperparameter optimisation with the tangent linear model}
The mode-corrected evidence approximation $\gL_h( \gM; w_t)$ matches the exact evidence of a \emph{tangent model}  with regressor $h: \R^{d_w} \times \R^{d_x} \to \R$ given by a first-order Taylor approximation to $f$ at $w_t$
\begin{gather*}
y = h(v) + \epsilon \;\; \textrm{with} \;\; h(v) \coloneqq f(w_t) + J (v - w_t) \;\; \textrm{and} \;\;  v \sim \gN(0, \alpha^{-1} I_{d_w}), \;\; \epsilon \sim \gN(0, \beta^{-1} I_n).
\end{gather*} 
The tangent model's loss is $\ell_h(v; \gM) = \frac{\beta}{2} \|y - h(v)\|^2 + \frac{\alpha}{2} \|v\|^2$. It is minimised by the MAP $v^\star \coloneqq \argmin_v \ell_h(v; \gM)$, which matches the optimum of the quadratic expansion in \cref{eq:corrected_laplace_evidence}, and its constant curvature matches the GGN $\partial^2_v \ell_h(v^\star; \gM) = H$. Thus, the above described mode-corrected algorithm selects hyperparameters that improve the tangent model's evidence $\gL_h( \gM; w_t)$ at each step. We hereon refer to this procedure as online Linear Model (LM).

\paragraph{Online Laplace as a variational bound on the tangent model's evidence}

Finally, we connect the LM procedure to the OL procedure described in \cref{subsec:preliminaries_online}. For this, we apply \cite{antoran2023samplingbased}'s lower bound on the linear model's evidence, i.e., the ELBO,
\begin{align}\label{eq:amazing_bound}
    &\gL_h( \gM; w_t) \geq \E_{\gN(v ; \, \mu, H^{-1})}[\log p_h(y,v; \gM)] + \gH(\gN(\mu, H^{-1})) \coloneqq \gL( \gM; \mu, w_t) \\
    &= \frac{d_w}{2} \log \alpha
    + \frac{n}{2} \log \beta
    - \frac{\beta}{2} \lVert y - (f(w_t) + J(\mu - w_t)) \rVert_2^2
    - \frac{\alpha}{2} \lVert \mu \rVert_2^2
    - \frac{1}{2}\log \det H
    - \frac{n}{2} \log 2\pi \notag
\end{align}
where $\gH$ refers to the differential entropy and $p_h(y, v; \gM) \coloneqq e^{-\ell_h(v; \gM)} \left(\frac{\beta}{2 \pi} \right)^{\frac{n}{2}} \left(\frac{\alpha}{2 \pi}\right)^{\frac{d_w}{2}} $.

The bound is tight when the variational posterior mean $\mu$ matches the linear model's MAP $v^\star$, that is, $\gL( \gM; v^\star, w_t) = \gL_h( \gM; w_t)$. This recovers the LM objective. On the other hand, by choosing the variational mean to match the linearisation point, $\mu = w_t$, we recover the OL evidence $\gL(\gM; w_t, w_t) = \gL_f(\gM; w_t)$. Thus, discarding the first-order term in the Taylor expansion of $\log p_f(y, w; \gM)$, which seemed unjustified, actually results in a lower bound on a variant of the Laplace evidence which corrects for the expansion point not matching the mode, that is, $\gL_f(\gM; w_t) \leq \gL_h(\gM; w_t)$.
This is illustrated in~\cref{fig:ELBO_alpha_trace}~(left).

\begin{figure}[t]
\vspace{-6pt}
\centering
\includegraphics[width=\textwidth]{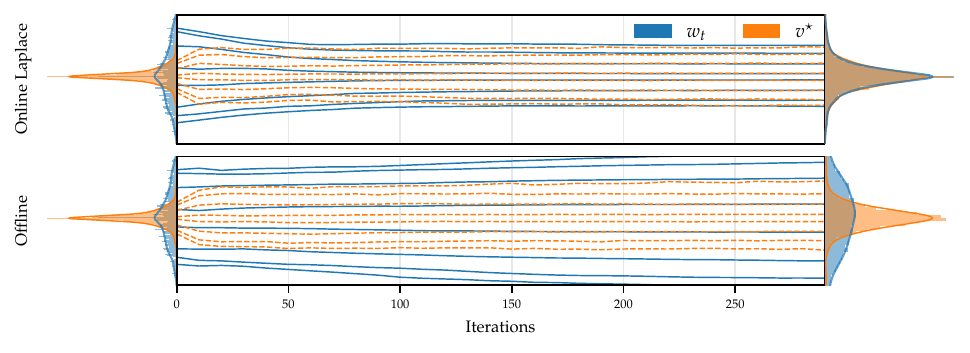}
\vspace{-6pt}
\caption{Evolution of neural network weights $w_t$ and tangent linear model posterior mean $v^\star$ during training.
With online Laplace (OL procedure), $w_t$ and $v^\star$ converge to the same distribution. Offline training does not exhibit this behaviour.}
\label{fig:main_histogram_convergence}
\vspace{-10pt}
\end{figure}

\paragraph{Convergence behaviour}

Both the OL and LM procedures reach a shared fixed point when a simultaneous optimum of the NN weights and linear model hyperparameters is found, that is, $w^\star \,{=}\, \argmin_w \ell_f(w; \gM^\star)$ and $\gM^\star \! = \! \argmax_\gM \gL_h(\gM; w^\star) =  \argmax_\gM \gL(\gM; w^\star, w^\star)$. Since the first-order term vanishes here, $\partial_w \ell_f(w^\star; \gM) \,{=}\, 0$, the ELBO is tight and the linear model evidence $\gL_h$ matches the regular Laplace evidence $\gL_f$, satisfying the termination condition of \cite{Mackay1992Thesis}'s NN re-training algorithm. 

At convergence, we have $\partial_w \ell_f(w^\star; \gM) \,{=}\, \partial_v \ell_h(w^\star; \gM) \,{=}\, 0$. We may thus use similarity between $w^\star$ and $v^\star$ to check for convergence of the OL and LM procedures. \cref{fig:main_histogram_convergence} shows the LM procedure results in the NN and tangent model converging to a shared MAP. This does not occur when performing standard NN training with fixed~hyperparameters. 

\begin{figure}[b]
\vspace{-6pt}
\centering
\includegraphics[width=0.98\textwidth]{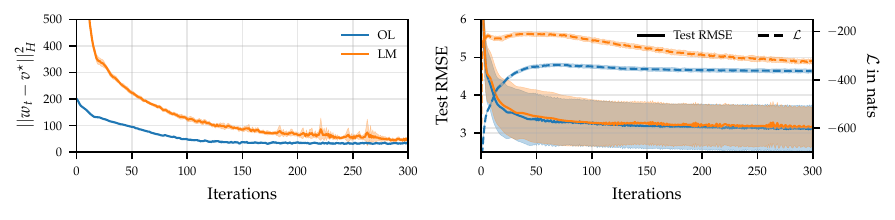}
\vspace{-5pt}
\caption{Difference between neural network weights $w_t$ and linear model posterior mean $v^\star$ (left) and test RMSE versus ELBO $\gL$ (right) throughout online training for both OL and LM procedures.
A maximised ELBO does not imply optimal test RMSE.}
\label{fig:metric_trace}
\vspace{-8pt}
\end{figure}

Since the NN and linear model share hyperparameters $\gM$, their loss gradients match at the linearisation point $\partial_w \ell_f(w_t; \gM) = \partial_v \ell_h(w_t;( \gM)$. Thus, during training, each NN optimisation step brings its parameters closer to the MAP of the current tangent model while also changing the linearisation point, which in turn changes the tangent model. This is not guaranteed to result in a tangent linear model with increased evidence. Indeed, \cref{fig:metric_trace} (right) shows both the OL and LM evidences exhibit non-monotonic training curves. Interestingly, the slack in the OL bound results in smaller hyperparameter steps, as shown in \cref{fig:ELBO_alpha_trace}. These may counteract training instability coming from a rapidly changing tangent model evidence. We find OL updates often lead to more stable convergence to a shared MAP for the NN and tangent model than LM updates. This can be seen in \cref{fig:metric_trace} (left).

\section{Experiments}

\begin{figure}[t]
\vspace{-6pt}
\centering
\includegraphics[width=0.98\textwidth]{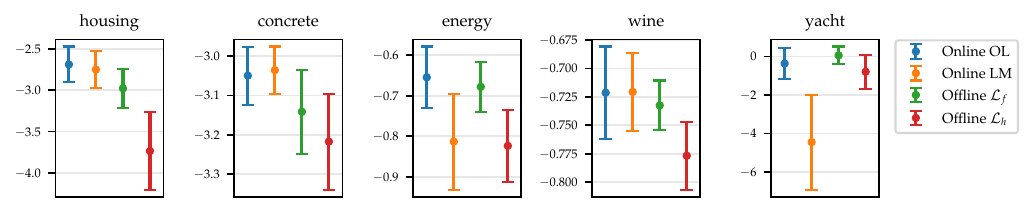}
\vspace{-3pt}
\caption{Test log-likelihood on UCI regression (mean $\pm$ standard error over 10 splits).}
\label{fig:llh_test_main}
\vspace{-7pt}
\end{figure}

This section demonstrates the OL and LM online hyperparameter tuning algorithms discussed in the previous section and contrasts them with standard NN training, that is, with fixed hyperparameters (labelled offline). To this end, we train single hidden layer, 50 hidden unit MLPs on 5 UCI datasets \citep{UCI_repo}: housing, concrete, energy, wine, and yacht. 
We train our NNs with full-batch Adam and an exponentially decaying learning rate. For online methods, we train until convergence, updating the hyperparameters after every NN parameter step. The hyperparameter update step is detailed in \cref{app:mackay_update}. For offline methods, we select the stopping point that minimises RMSE on a validation set consisting of 10\% of training data points. We do not use normalisation layers, as these interfere with the Laplace approximation \citep{antoran2022linearised}.

\paragraph{Convergence of NN and linear model posteriors} \cref{fig:metric_trace} (left) shows the Euclidean distance between the NN parameters and tangent model MAP decreasing throughout training for both the LM and OL procedures. Although the distance does not exactly reach 0, the NN and linear model parameter vectors present nearly identical distributions, as shown in \cref{fig:main_histogram_convergence}. This does not occur in the offline setting. The LM procedure takes larger hyperparameter steps than OL resulting in some training instability and overall slower convergence, as shown in \cref{fig:ELBO_alpha_trace}. \cref{fig:metric_trace} (right) shows that both the LM and OL evidence estimates attain their maxima very early in training, before the lowest test RMSE is reached. As discussed in \cref{sec:contributions}, these metrics target the evidence of a tangent model and may only transfer to the NN once online training has fully converged.

\paragraph{Predictive performance}

We obtain predictive distributions using the linearised Laplace method \citep{Immer2021Improving}. For NNs trained offline, linearised Laplace hyperparameters are chosen post-hoc my maximising either the $\gL_f$ or $\gL_h$ objectives. \cref{fig:llh_test_main} reports log-likelihood mean and standard errors obtained across 10 train-test splits where the test set size is 10\% of the whole dataset size. 

We find the OL procedure to consistently provide the best performance or be a close second best. The LM procedure performs very similarly to OL on all datasets except energy and yacht, where it performs the worse out of all methods. We attribute this to training instability. Offline methods only perform competitively on yacht, despite use of validation-based stopping. The online methods' improvement over their offline counterparts can mostly be explained by improved RMSE (see \cref{fig:rmse_test}), suggesting online hyperparameter optimisation helps find NN parameters that effectively navigate  the bias-variance trade off.

The full set of experimental results, including hyperparameter trajectories and evidence estimates, is provided in \cref{app:more_results}.


\clearpage



    



    

\section*{Acknowledgements}
JAL was supported by the University of Cambridge Harding Distinguished Postgraduate Scholars Programme. JA acknowledges support from Microsoft Research, through its PhD Scholarship Programme, and from the EPSRC. JMHL acknowledges support from a Turing AI Fellowship under grant EP/V023756/1.

\bibliography{references}

\begin{thebibliography}{26}
\providecommand{\natexlab}[1]{#1}
\providecommand{\url}[1]{\texttt{#1}}
\expandafter\ifx\csname urlstyle\endcsname\relax
  \providecommand{\doi}[1]{doi: #1}\else
  \providecommand{\doi}{doi: \begingroup \urlstyle{rm}\Url}\fi

\bibitem[Antor{\'{a}}n et~al.(2022{\natexlab{a}})Antor{\'{a}}n, Allingham,
  Janz, Daxberger, Nalisnick, and Hern{\'a}ndez-Lobato]{antoran2022linearised}
Javier Antor{\'{a}}n, James~Urquhart Allingham, David Janz, Erik Daxberger,
  Eric Nalisnick, and Jos{\'e}~Miguel Hern{\'a}ndez-Lobato.
\newblock {L}inearised {L}aplace {I}nference in {N}etworks with {N}ormalisation
  {L}ayers and the {N}eural g-{P}rior.
\newblock In \emph{Advances in Approximate Bayesian Inference},
  2022{\natexlab{a}}.

\bibitem[Antor{\'{a}}n et~al.(2022{\natexlab{b}})Antor{\'{a}}n, Janz,
  Allingham, Daxberger, Barbano, Nalisnick, and
  Hern{\'{a}}ndez{-}Lobato]{antoran2022Adapting}
Javier Antor{\'{a}}n, David Janz, James~Urquhart Allingham, Erik~A. Daxberger,
  Riccardo Barbano, Eric~T. Nalisnick, and Jos{\'{e}}~Miguel
  Hern{\'{a}}ndez{-}Lobato.
\newblock {A}dapting the {L}inearised {L}aplace {M}odel {E}vidence for {M}odern
  {D}eep {L}earning.
\newblock In \emph{International Conference on Machine Learning}, volume 162,
  pages 796--821, 2022{\natexlab{b}}.

\bibitem[Antorán et~al.(2023)Antorán, Padhy, Barbano, Nalisnick, Janz, and
  Hern{\'a}ndez-Lobato]{antoran2023samplingbased}
Javier Antorán, Shreyas Padhy, Riccardo Barbano, Eric Nalisnick, David Janz,
  and Jos{\'e}~Miguel Hern{\'a}ndez-Lobato.
\newblock {S}ampling-based inference for large linear models, with application
  to linearised {L}aplace.
\newblock In \emph{International Conference on Learning Representations}, 2023.

\bibitem[Barbano et~al.(2023)Barbano, Antor{\'{a}}n, Leuschner,
  Hern{\'{a}}ndez{-}Lobato, Kereta, and Jin]{barbano2023fast}
Riccardo Barbano, Javier Antor{\'{a}}n, Johannes Leuschner, Jos{\'{e}}~Miguel
  Hern{\'{a}}ndez{-}Lobato, Zeljko Kereta, and Bangti Jin.
\newblock {F}ast and {P}ainless {I}mage {R}econstruction in {D}eep {I}mage
  {P}rior {S}ubspaces.
\newblock In \emph{arXiv abs/2302.10279}, 2023.

\bibitem[Bishop(2006)]{bishop2006pattern}
Christopher~M Bishop.
\newblock \emph{{P}attern {R}ecognition and {M}achine {L}earning}.
\newblock springer, 2006.

\bibitem[Botev et~al.(2017)Botev, Ritter, and Barber]{botev2017Optimisation}
Aleksandar Botev, Hippolyt Ritter, and David Barber.
\newblock {P}ractical {G}auss-{N}ewton {O}ptimisation for {D}eep {L}earning.
\newblock In \emph{International Conference on Machine Learning}, 2017.

\bibitem[Clarke et~al.(2022)Clarke, Oldewage, and
  Hern{\'a}ndez-Lobato]{clarke2022scalable}
Ross~M Clarke, Elre~Talea Oldewage, and Jos{\'e}~Miguel Hern{\'a}ndez-Lobato.
\newblock {S}calable {O}ne-{P}ass {O}ptimisation of {H}igh-{D}imensional
  {W}eight-{U}pdate {H}yperparameters by {I}mplicit {D}ifferentiation.
\newblock In \emph{International Conference on Learning Representations}, 2022.

\bibitem[Dua and Graff(2017)]{UCI_repo}
Dheeru Dua and Casey Graff.
\newblock {UCI} {M}achine {L}earning {R}epository, 2017.

\bibitem[Foresee and Hagan(1997)]{Foresee97BayesNewton}
F.~Dan Foresee and Martin~T. Hagan.
\newblock {G}auss-{N}ewton approximation to {B}ayesian learning.
\newblock In \emph{International Conference on Neural Networks}, 1997.

\bibitem[Friston et~al.(2007)Friston, Mattout, Trujillo{-}Barreto, Ashburner,
  and Penny]{Friston07variational}
Karl~J. Friston, J{\'{e}}r{\'{e}}mie Mattout, Nelson~J. Trujillo{-}Barreto,
  John Ashburner, and William~D. Penny.
\newblock {V}ariational free energy and the {L}aplace approximation.
\newblock \emph{NeuroImage}, 34\penalty0 (1):\penalty0 220--234, 2007.

\bibitem[Gull(1989)]{Gull1989line}
Stephen~F. Gull.
\newblock \emph{{B}ayesian {D}ata {A}nalysis: {S}traight-line {f}itting}, pages
  511--518.
\newblock Springer Netherlands, 1989.

\bibitem[Immer et~al.(2021{\natexlab{a}})Immer, Bauer, Fortuin, R{\"{a}}tsch,
  and Khan]{Immer2021Marginal}
Alexander Immer, Matthias Bauer, Vincent Fortuin, Gunnar R{\"{a}}tsch, and
  Mohammad~Emtiyaz Khan.
\newblock {S}calable {M}arginal {L}ikelihood {E}stimation for {M}odel
  {S}election in {D}eep {L}earning.
\newblock In \emph{International Conference on Machine Learning},
  2021{\natexlab{a}}.

\bibitem[Immer et~al.(2021{\natexlab{b}})Immer, Korzepa, and
  Bauer]{Immer2021Improving}
Alexander Immer, Maciej Korzepa, and Matthias Bauer.
\newblock Improving predictions of {B}ayesian neural nets via local
  linearization.
\newblock In \emph{International Conference on Artificial Intelligence and
  Statistics}, 2021{\natexlab{b}}.

\bibitem[Immer et~al.(2022)Immer, van~der Ouderaa, Ratsch, Fortuin, and van~der
  Wilk]{immer2022invariance}
Alexander Immer, Tycho~F.A. van~der Ouderaa, Gunnar Ratsch, Vincent Fortuin,
  and Mark van~der Wilk.
\newblock {I}nvariance {L}earning in {D}eep {N}eural {N}etworks with
  {D}ifferentiable {L}aplace {A}pproximations.
\newblock In \emph{Advances in Neural Information Processing Systems}, 2022.

\bibitem[Immer et~al.(2023)Immer, van~der Ouderaa, van~der Wilk, R{\"{a}}tsch,
  and Sch{\"{o}}lkopf]{Immer2023NTK}
Alexander Immer, Tycho F.~A. van~der Ouderaa, Mark van~der Wilk, Gunnar
  R{\"{a}}tsch, and Bernhard Sch{\"{o}}lkopf.
\newblock {S}tochastic {M}arginal {L}ikelihood {G}radients using {N}eural
  {T}angent {K}ernels.
\newblock In \emph{arXiv abs/2306.03968}, 2023.

\bibitem[Khan et~al.(2019)Khan, Immer, Abedi, and Korzepa]{Khan2019Approximate}
Mohammad~Emtiyaz Khan, Alexander Immer, Ehsan Abedi, and Maciej Korzepa.
\newblock {A}pproximate {I}nference {T}urns {D}eep {N}etworks into {G}aussian
  {P}rocesses.
\newblock In \emph{Advances in Neural Information Processing Systems}, 2019.

\bibitem[Lin et~al.(2023)Lin, Antorán, Padhy, Janz, Hernández-Lobato, and
  Terenin]{lin2023sampling}
Jihao~Andreas Lin, Javier Antorán, Shreyas Padhy, David Janz, José~Miguel
  Hernández-Lobato, and Alexander Terenin.
\newblock {S}ampling from {G}aussian {P}rocess {P}osteriors using {S}tochastic
  {G}radient {D}escent, 2023.

\bibitem[Mackay(1992)]{Mackay1992Thesis}
David J.~C. Mackay.
\newblock \emph{Bayesian Methods for Adaptive Models}.
\newblock PhD thesis, 1992.

\bibitem[Martens(2020)]{Martens2020insights}
James Martens.
\newblock {N}ew {I}nsights and {P}erspectives on the {N}atural {G}radient
  {M}ethod.
\newblock \emph{Journal of Machine Learning Research}, 21:\penalty0
  146:1--146:76, 2020.

\bibitem[Rasmussen and Williams(2006)]{RasmussenW06}
Carl~Edward Rasmussen and Christopher K.~I. Williams.
\newblock \emph{{G}aussian {P}rocesses for {M}achine {L}earning}.
\newblock MIT Press, 2006.

\bibitem[Ritter et~al.(2018)Ritter, Botev, and Barber]{ritter2018scalable}
Hippolyt Ritter, Aleksandar Botev, and David Barber.
\newblock {A} {S}calable {L}aplace {A}pproximation for {N}eural {N}etworks.
\newblock In \emph{International Conference on Learning Representations}, 2018.

\bibitem[Tipping(2001)]{Tipping2001sparse}
Michael~E. Tipping.
\newblock {S}parse {B}ayesian {L}earning and the {R}elevance {V}ector
  {M}achine.
\newblock \emph{Journal of Machine Learning Research}, 1:\penalty0 211--244,
  2001.

\bibitem[Tipping and Faul(2003)]{Tipping2003fast}
Michael~E. Tipping and Anita~C. Faul.
\newblock {F}ast {M}arginal {L}ikelihood {M}aximisation for {S}parse {B}ayesian
  {M}odels.
\newblock In \emph{International Conference on Artificial Intelligence and
  Statistics}, 2003.

\bibitem[Watson et~al.(2020)Watson, Lin, Klink, and Peters]{watson2020neural}
Joe Watson, Jihao~Andreas Lin, Pascal Klink, and Jan Peters.
\newblock {N}eural {L}inear {M}odels with {F}unctional {G}aussian {P}rocess
  {P}riors.
\newblock In \emph{Advances in Approximate Bayesian Inference}, 2020.

\bibitem[Watson et~al.(2021)Watson, Lin, Klink, Pajarinen, and
  Peters]{watson2021latent}
Joe Watson, Jihao~Andreas Lin, Pascal Klink, Joni Pajarinen, and Jan Peters.
\newblock {L}atent {D}erivative {B}ayesian {L}ast {L}ayer {N}etworks.
\newblock In \emph{International Conference on Artificial Intelligence and
  Statistics}, 2021.

\bibitem[Wipf and Nagarajan(2007)]{Wipf2007Determination}
David~P. Wipf and Srikantan~S. Nagarajan.
\newblock {A} {N}ew {V}iew of {A}utomatic {R}elevance {D}etermination.
\newblock In \emph{Advances in Neural Information Processing Systems}, 2007.

\end{thebibliography}
\clearpage

\appendix

\section{Evolution of neural network training} \label{app:more_results}

\begin{figure}[H]
\centering
\includegraphics[width=0.49\textwidth]{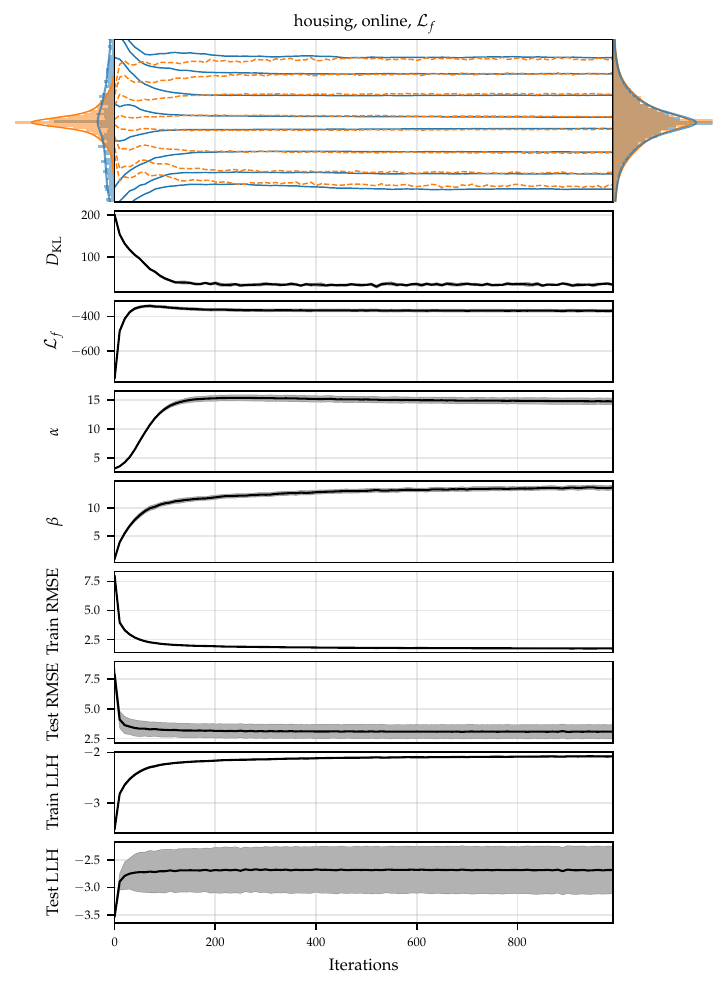}
\includegraphics[width=0.49\textwidth]{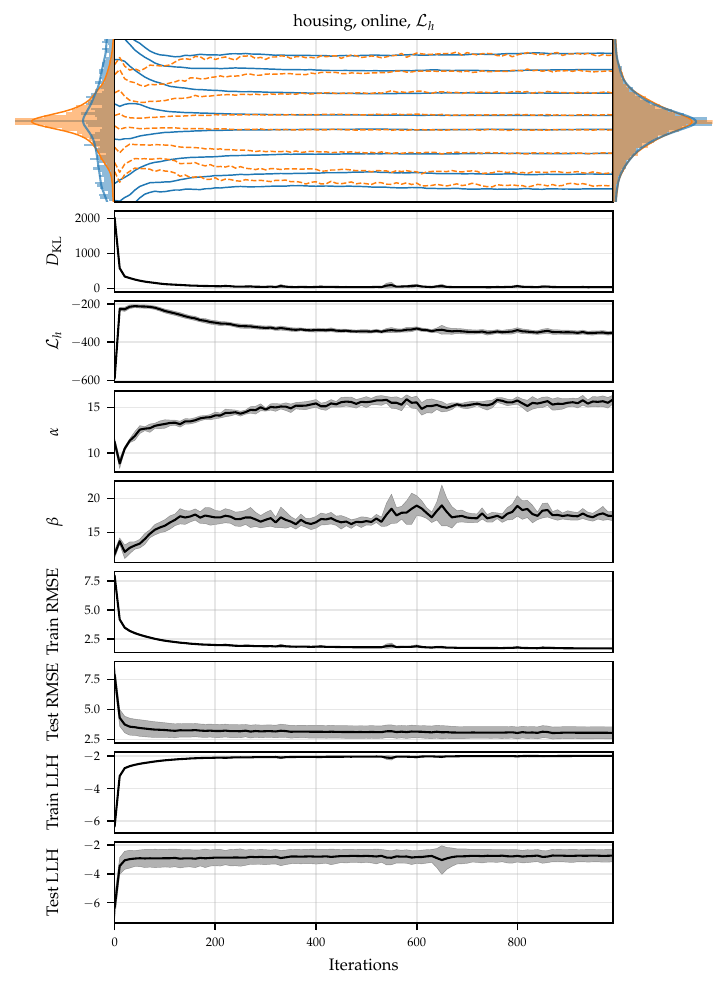} \\
\includegraphics[width=0.49\textwidth]{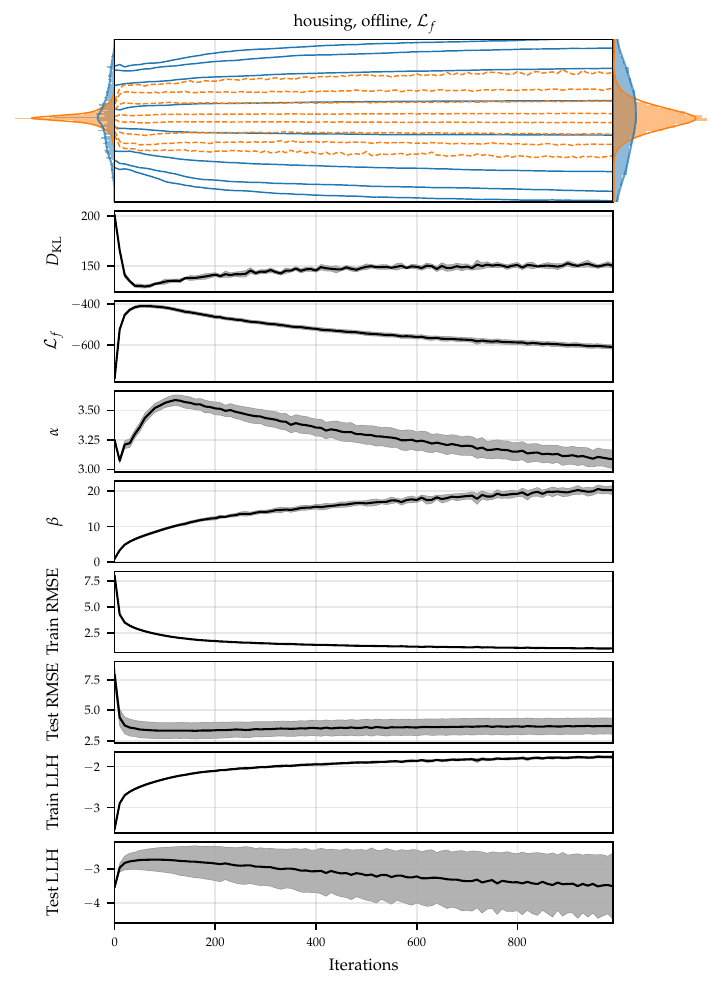}
\includegraphics[width=0.49\textwidth]{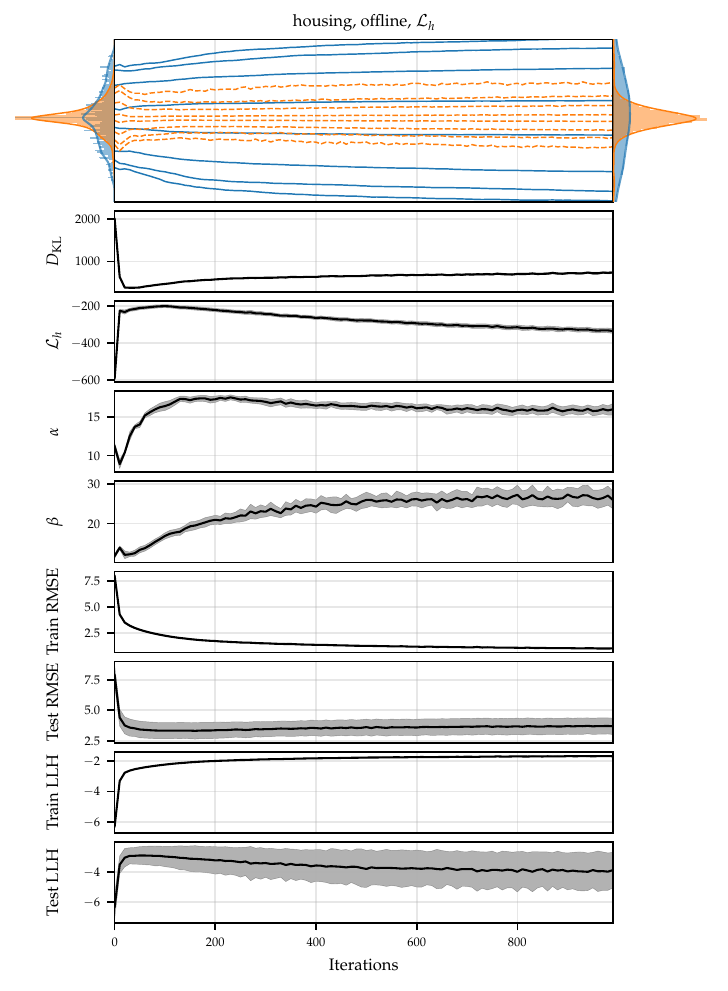}
\caption{Evolution of neural network training on the housing dataset.}
\label{fig:evolution_housing}
\end{figure}

\begin{figure}[H]
\centering
\includegraphics[width=0.49\textwidth]{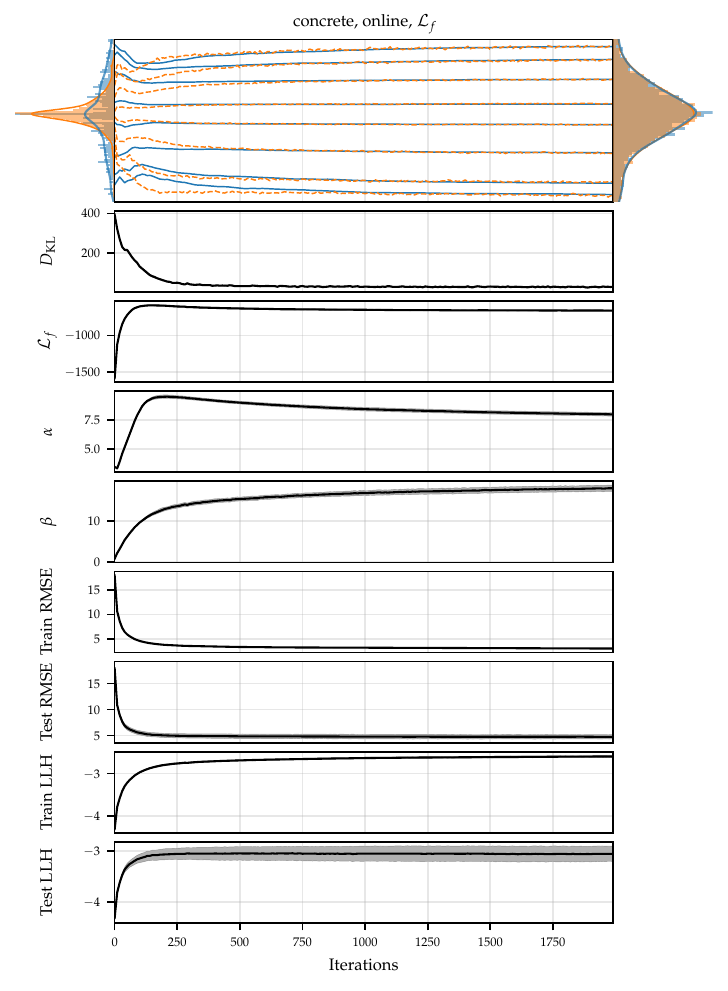}
\includegraphics[width=0.49\textwidth]{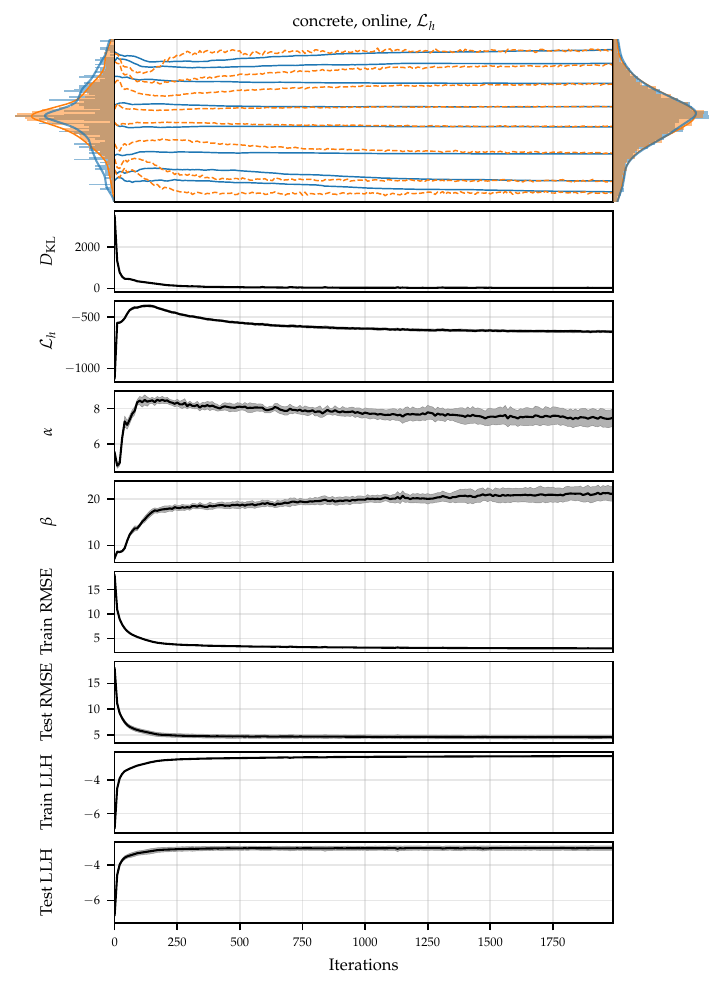} \\
\includegraphics[width=0.49\textwidth]{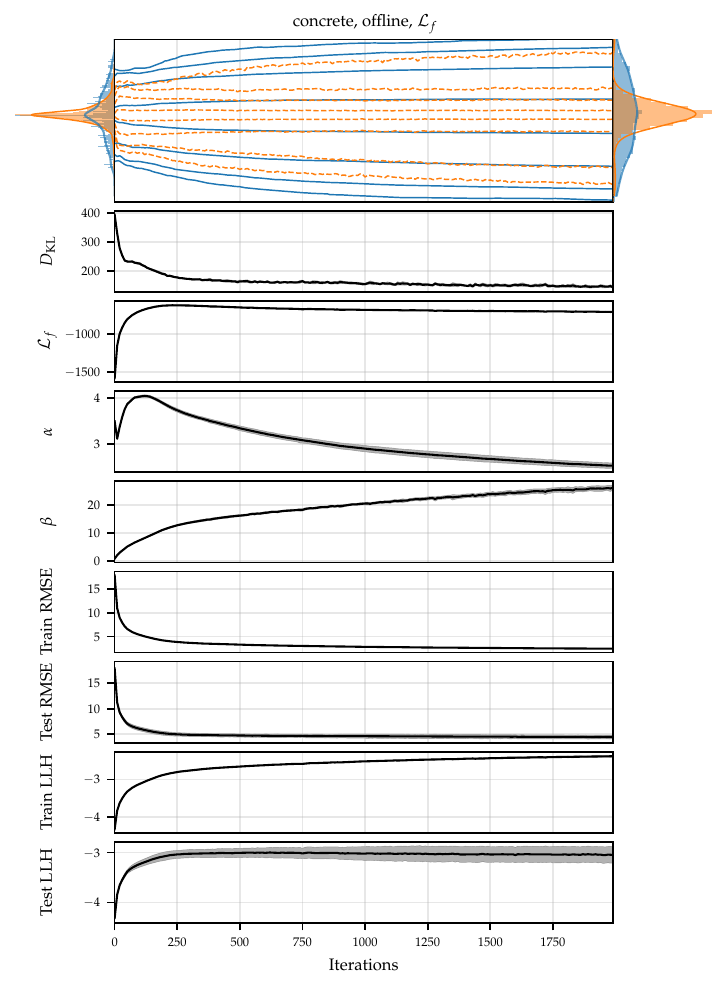}
\includegraphics[width=0.49\textwidth]{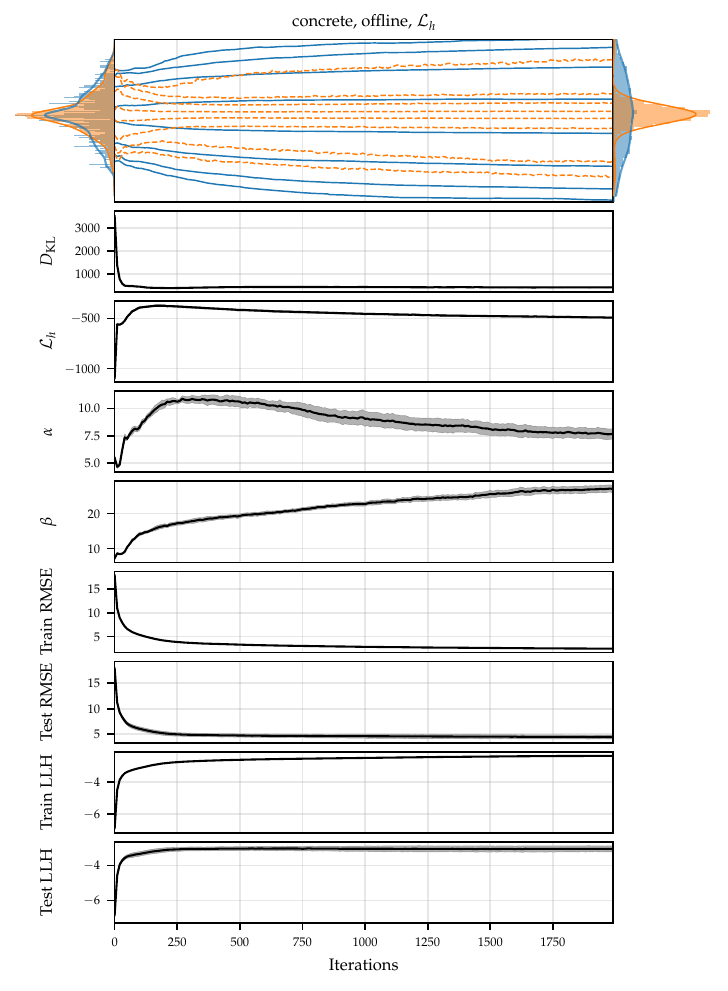}
\caption{Evolution of neural network training on the concrete dataset.}
\label{fig:evolution_concrete}
\end{figure}

\begin{figure}[H]
\centering
\includegraphics[width=0.49\textwidth]{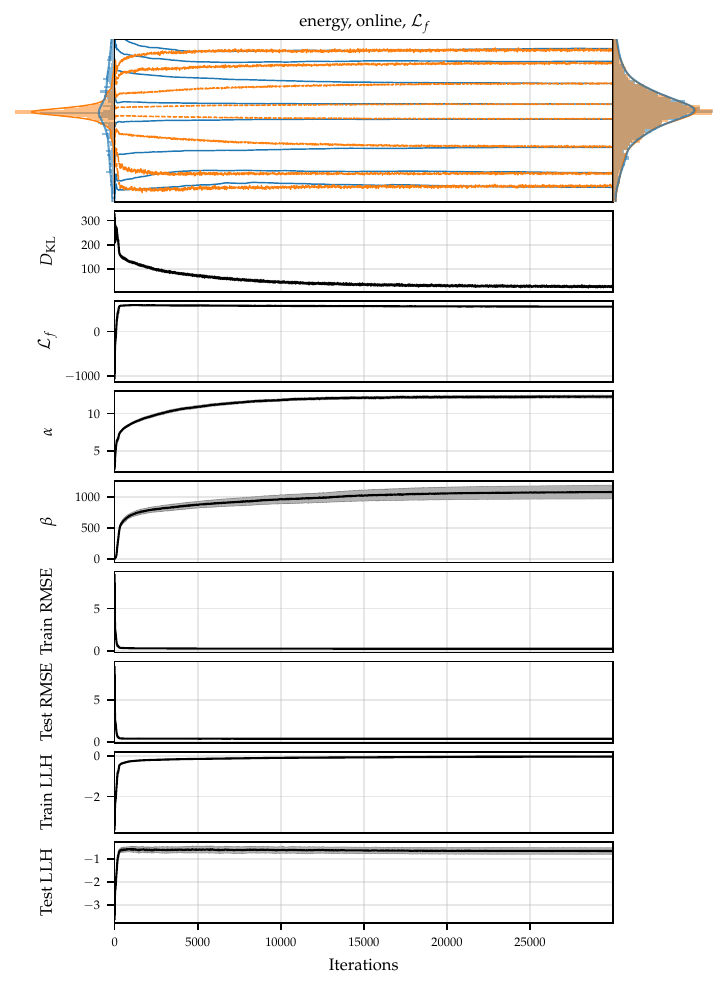}
\includegraphics[width=0.49\textwidth]{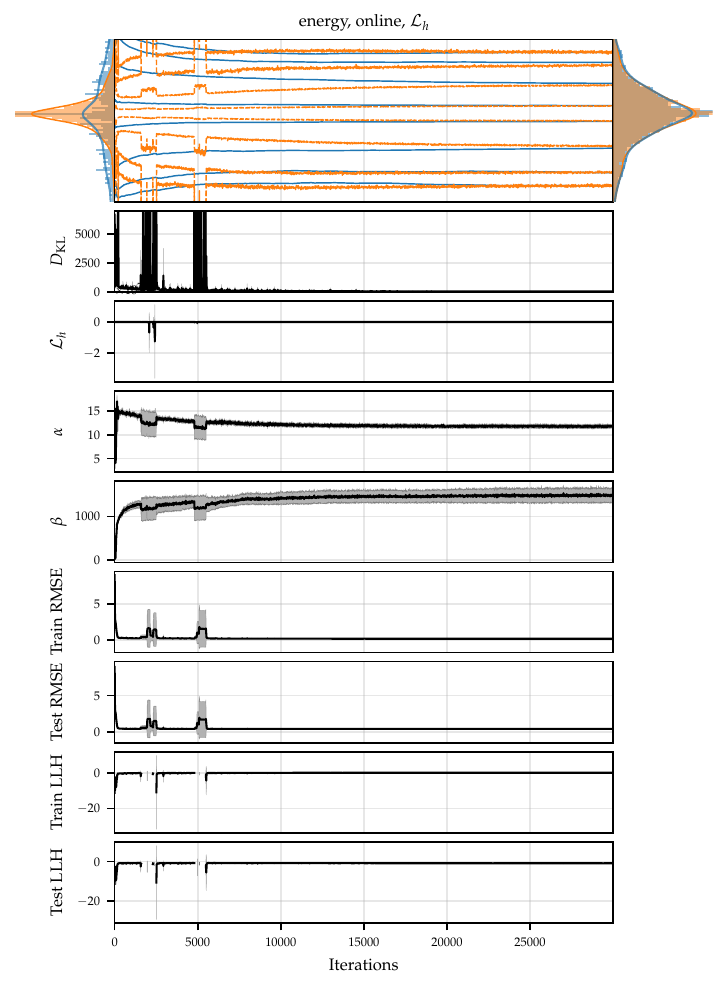} \\
\includegraphics[width=0.49\textwidth]{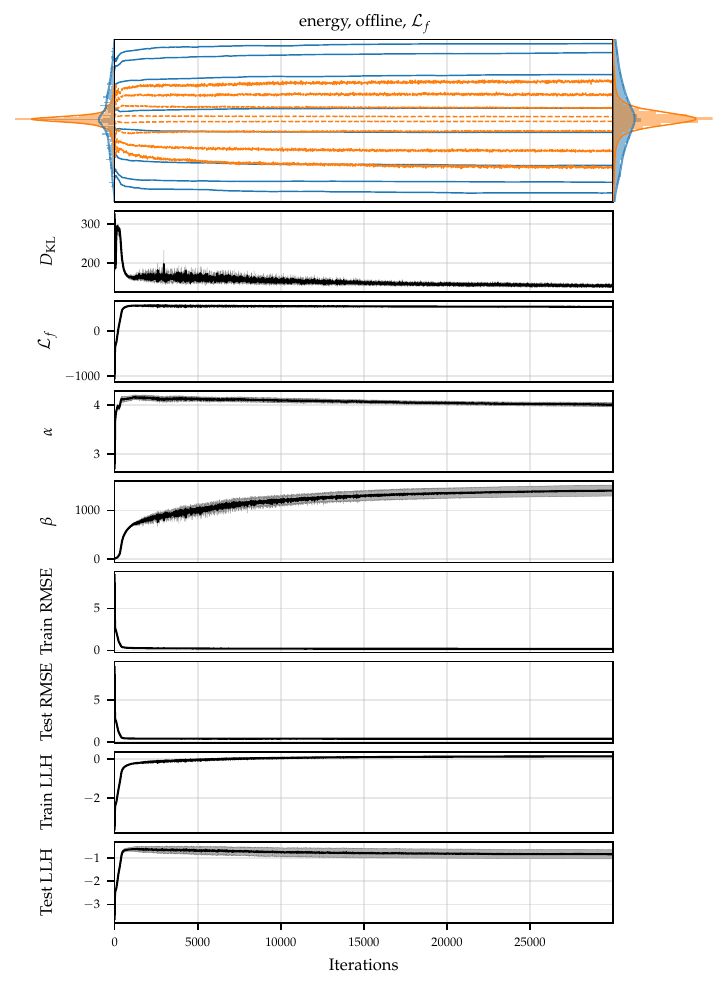}
\includegraphics[width=0.49\textwidth]{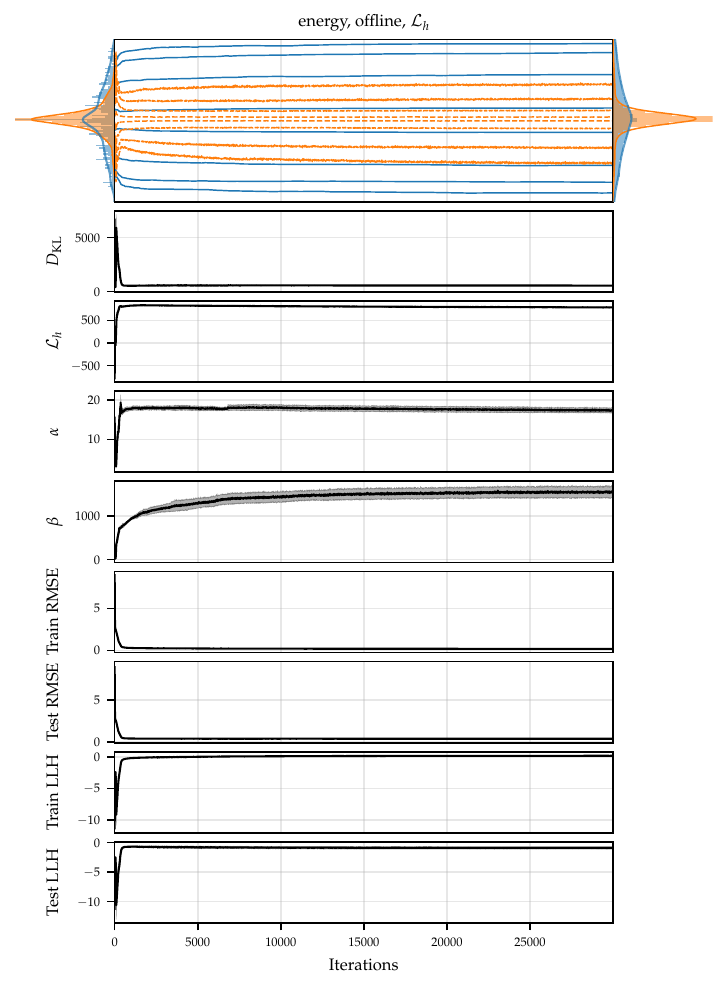}
\caption{Evolution of neural network training on the energy dataset.}
\label{fig:evolution_energy}
\end{figure}

\begin{figure}[H]
\centering
\includegraphics[width=0.49\textwidth]{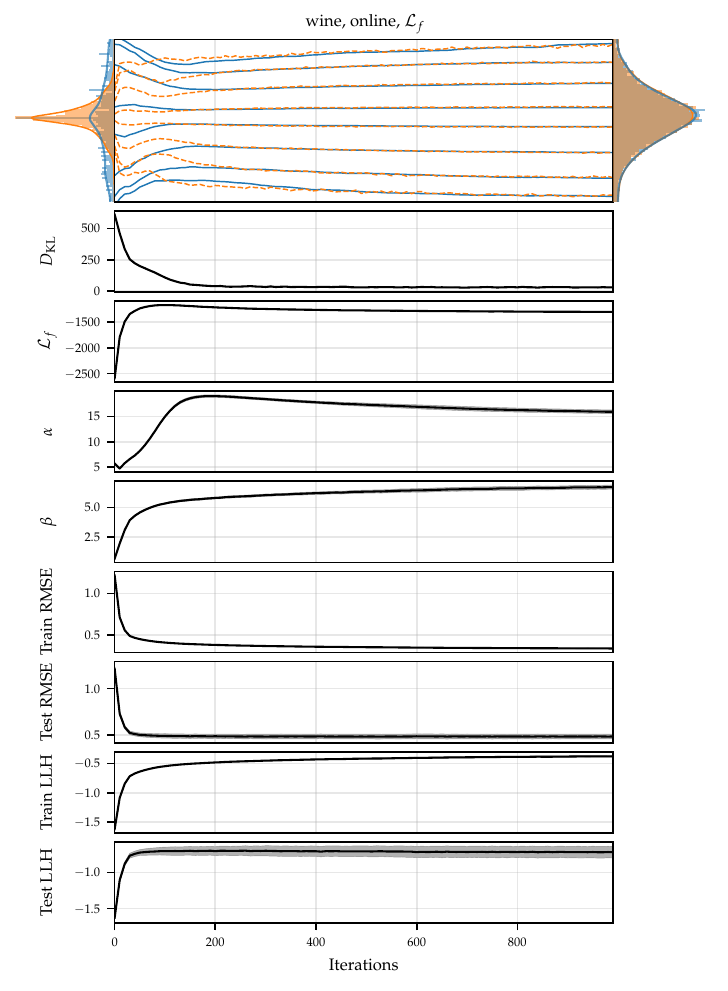}
\includegraphics[width=0.49\textwidth]{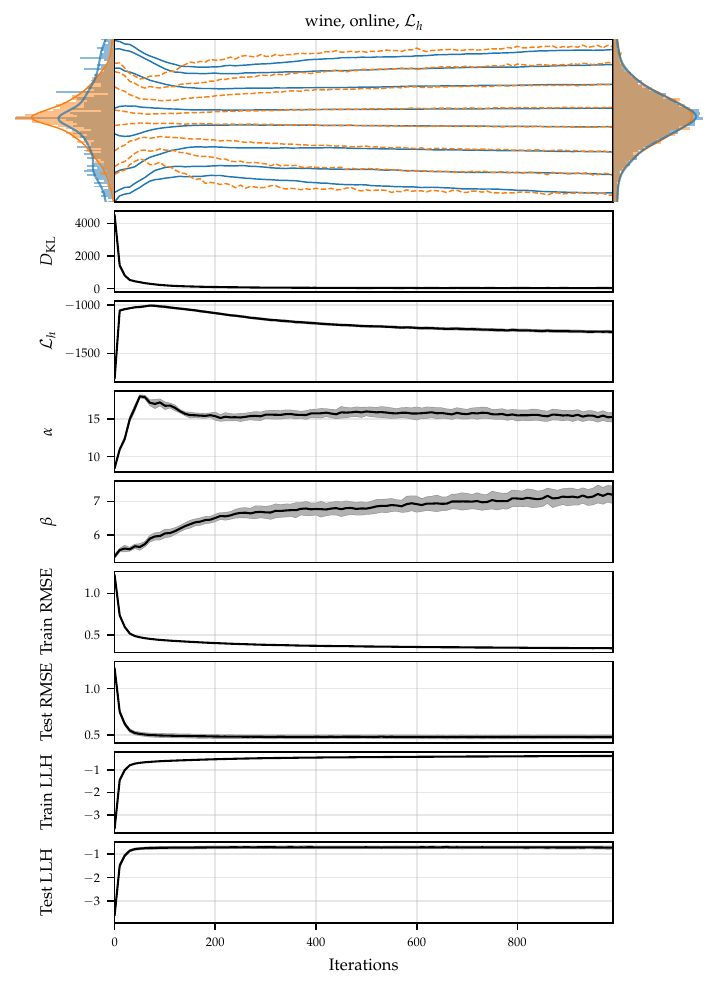} \\
\includegraphics[width=0.49\textwidth]{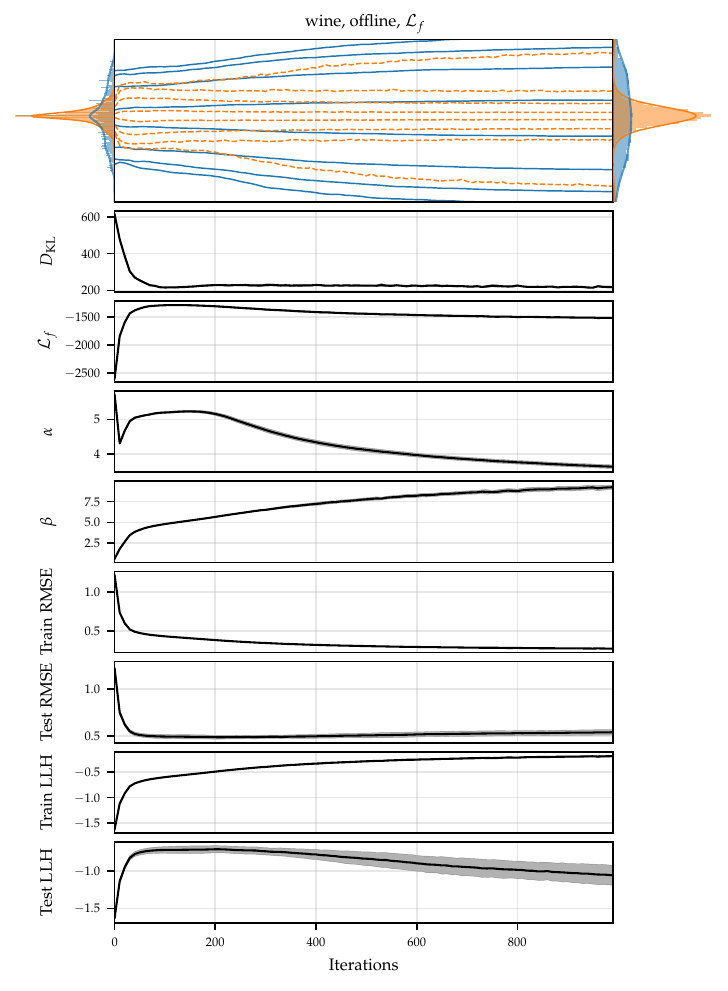}
\includegraphics[width=0.49\textwidth]{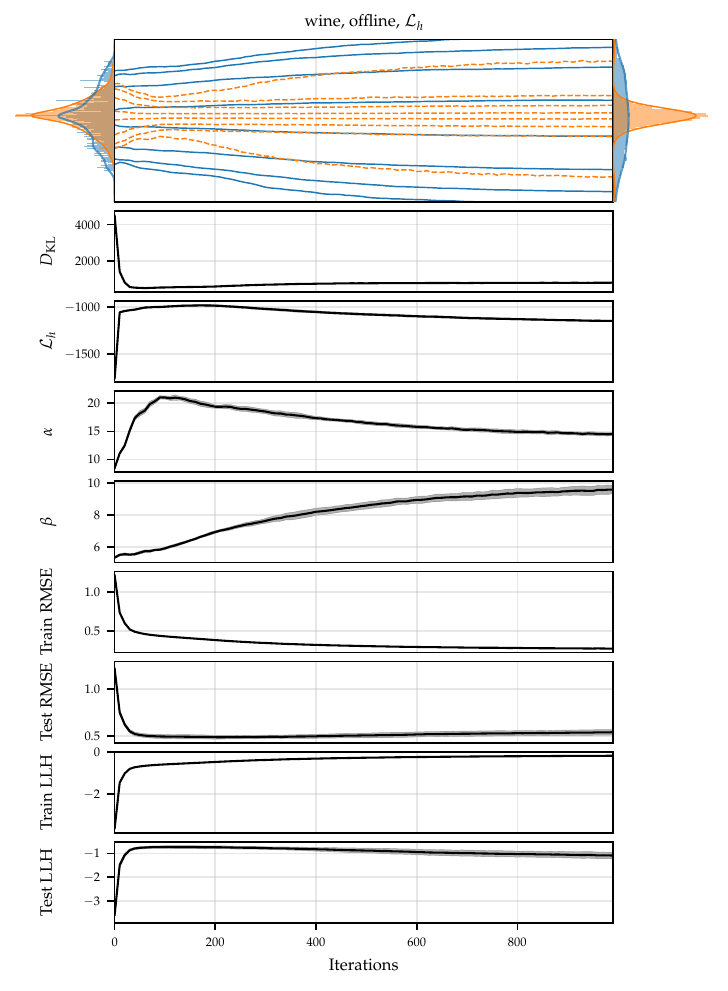}
\caption{Evolution of neural network training on the wine dataset.}
\label{fig:evolution_wine}
\end{figure}

\begin{figure}[H]
\centering
\includegraphics[width=0.49\textwidth]{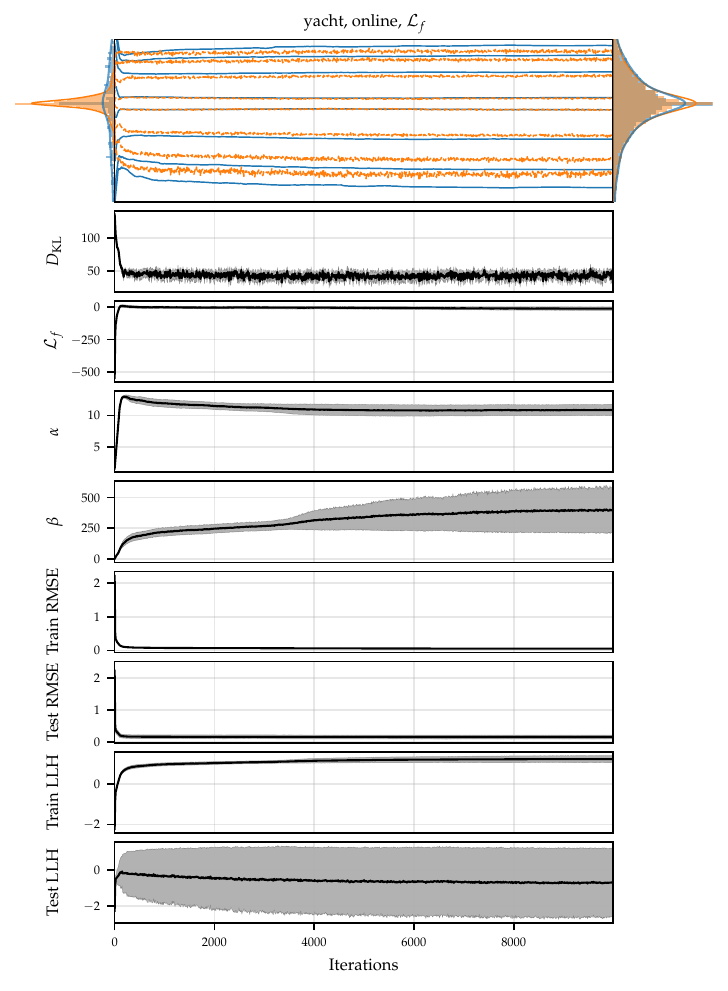}
\includegraphics[width=0.49\textwidth]{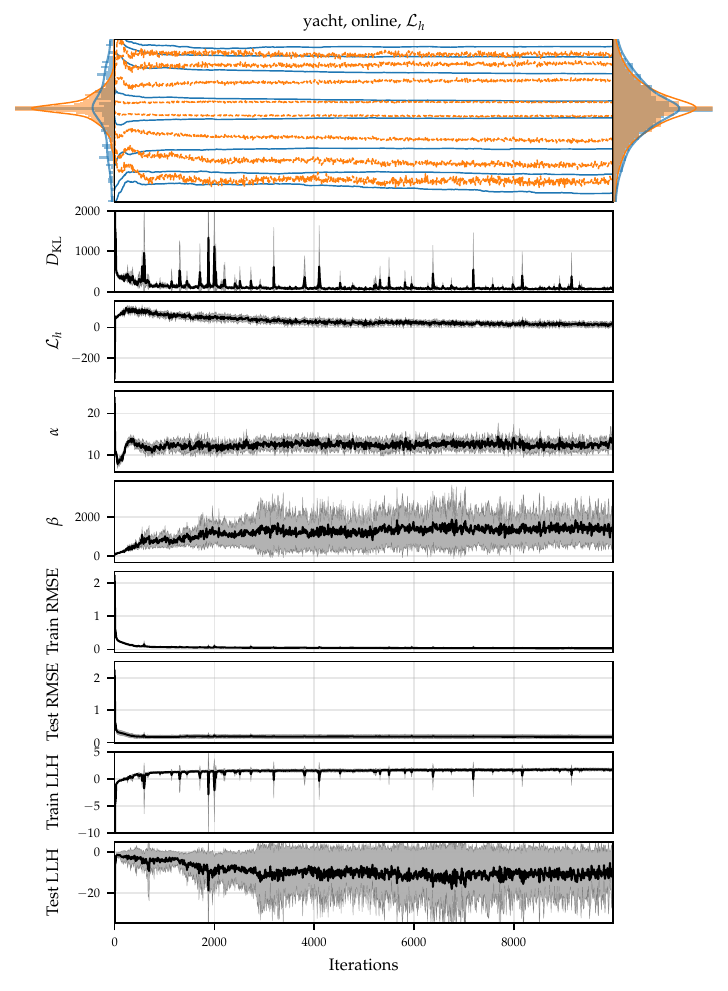} \\
\includegraphics[width=0.49\textwidth]{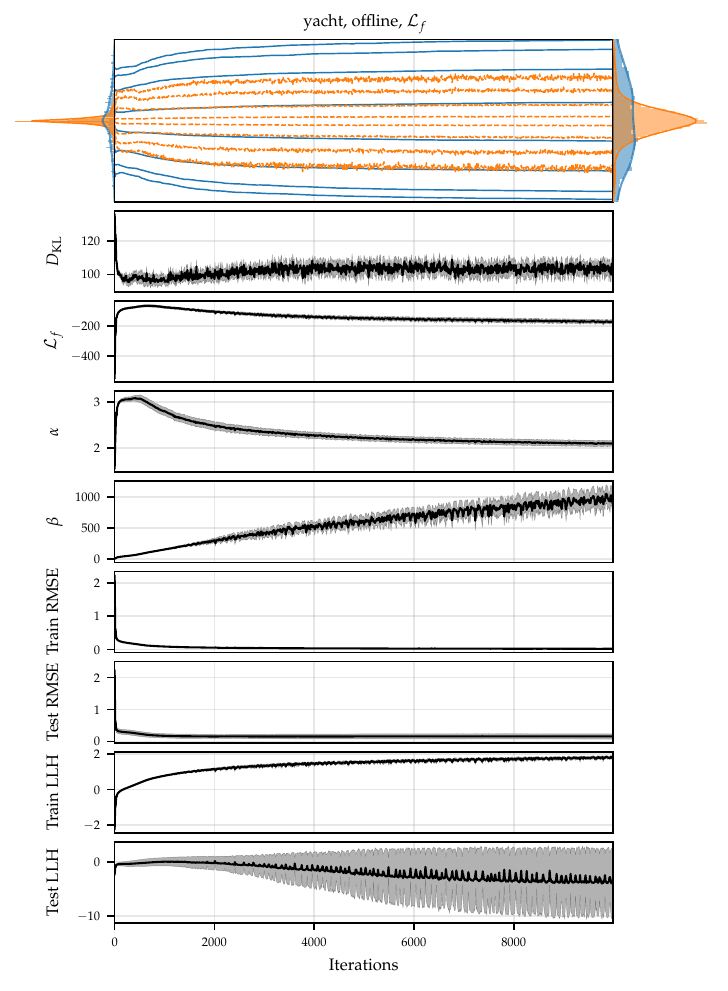}
\includegraphics[width=0.49\textwidth]{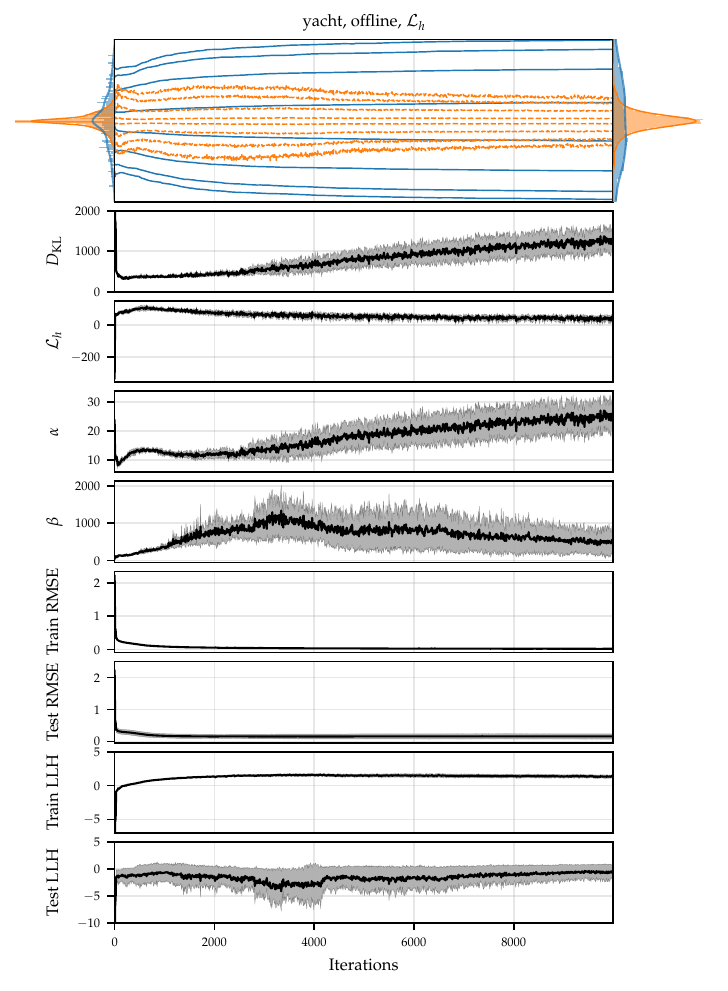}
\caption{Evolution of neural network training on the yacht dataset.}
\label{fig:evolution_yacht}
\end{figure}

\section{Additional predictive metrics}

\begin{figure}[H]
\centering
\includegraphics[width=0.98\textwidth]{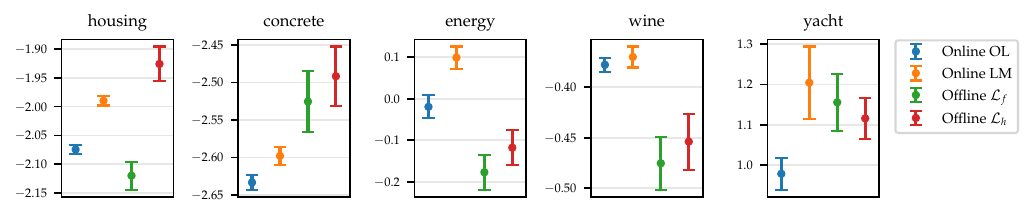}
\caption{Train log-likelihood on UCI regression (mean $\pm$ standard error over 10 splits).}
\label{fig:llh_train}
\end{figure}

\begin{figure}[H]
\centering
\includegraphics[width=0.98\textwidth]{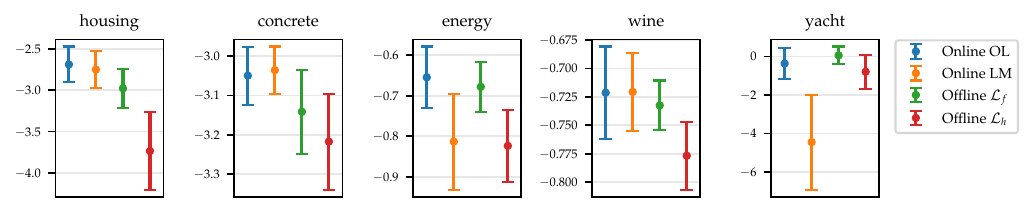}
\caption{Test log-likelihood on UCI regression (mean $\pm$ standard error over 10 splits).}
\label{fig:llh_test}
\end{figure}

\begin{figure}[H]
\centering
\includegraphics[width=0.98\textwidth]{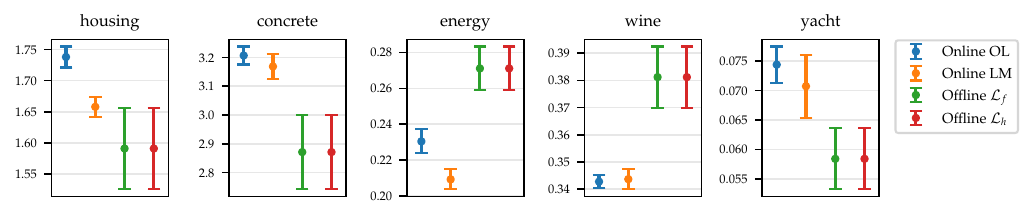}
\caption{Train RMSE on UCI regression (mean $\pm$ standard error over 10 splits).}
\label{fig:rmse_train}
\end{figure}

\begin{figure}[H]
\centering
\includegraphics[width=0.98\textwidth]{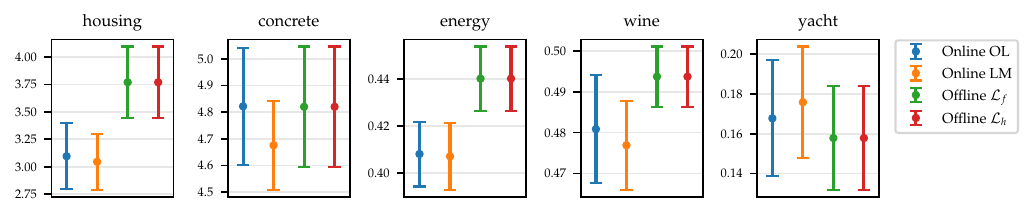}
\caption{Test RMSE on UCI regression (mean $\pm$ standard error over 10 splits).}
\label{fig:rmse_test}
\end{figure}
\newpage

\section{Related work}

This work builds on the rich literature of Bayesian linear regression, which was first introduced by \cite{Gull1989line}.
We recommend chapter 3 of \cite{bishop2006pattern} and chapter 2 of \cite{RasmussenW06} for an introduction. 
The use of both the linear model evidence and Laplace model evidence for hyperparameter selection are introduced in \cite{Mackay1992Thesis}. These techniques are analysed and extended by \cite{Tipping2001sparse,Tipping2003fast,Wipf2007Determination} for linear model and by \cite{Immer2021Marginal,antoran2022Adapting} for NNs.

Despite their analytic tractability, linear models are held back by a cost of inference cubic in the number of parameters when expressed in primal form, or cubic in the number of observations for the dual (i.e. kernelised or Gaussian Process) form. Although we do not deal with this computationally intractability in this work, there exist a number of approximations which aim to make inference in linearised NNs scalable \citep{ritter2018scalable,Khan2019Approximate,watson2020neural,watson2021latent,antoran2023samplingbased,Immer2023NTK,lin2023sampling}.

\section{MacKay's hyperparameter update}
\label{app:mackay_update}
Given mean $\mu$ and predictions $\hat{y}$, the hyperparameters $\alpha$ and $\beta$ can be updated by
\begin{align}
\label{eq:MacKay_update}
    \alpha &= \frac{\gamma}{\lVert \mu \rVert_2^2}, &
    \beta &= \frac{n - \gamma}{\lVert y - \hat{y} \rVert_2^2}, &
    \gamma &= d_w - \alpha \mathrm{Tr}(H^{-1}),
\end{align}
to maximise the linear model's evidence \citep{Mackay1992Thesis}, where $\gamma$ is the effective number of dimensions.
For OL, $\mu = w$ and $\hat{y} = f(w)$, and in the case of LM, $\mu = v^\star$ and $\hat{y} = h(v^\star)$.

\begin{figure}[H]
\centering
\includegraphics[width=0.99\textwidth]{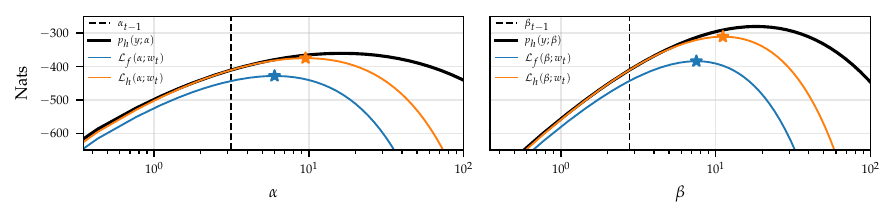}
\caption{Illustration of a hyperparameter update at a single timestep $t$.
Starting at the current value (dashed black), the updated value (star) is obtained by maximising the ELBO.}
\label{fig:ELBOs}
\end{figure}

\section{Implementation details}
\label{app:implementation_details}
The initial learning rate used for Adam was set to 0.01 and the exponential learning rate decay factor was set to 0.9999.
In the online setting, housing and wine were trained for 1000, concrete for 2000, energy for 30000 and yacht for 10000 steps until convergence.
In the offline setting, the validation RMSE was tracked for a maximum of 30000 epochs.

\end{document}